\documentclass[dvipsnames]{article}

    \usepackage[preprint]{neurips_2023}

\usepackage[utf8]{inputenc} 
\usepackage[T1]{fontenc}    
\usepackage{hyperref}       
\usepackage{url}            
\usepackage{booktabs}       
\usepackage{amsfonts}       
\usepackage{nicefrac}       
\usepackage{microtype}      
\usepackage{xcolor}         
\usepackage{graphicx}
\usepackage{subfigure}
\usepackage{booktabs} 
\usepackage{algorithm}
\usepackage[noend]{algorithmic}

\usepackage{amsmath}
\usepackage{amssymb}
\usepackage{mathtools}
\usepackage{amsthm}

\usepackage[capitalize,noabbrev]{cleveref}

\usepackage{url}
\usepackage{bbm}
\usepackage{booktabs} 
\usepackage{fixcmex}
\usepackage{bm}
\usepackage{mathtools}
\usepackage{multicol}
\usepackage{dcolumn}
\usepackage{tabularx}
\usepackage{paralist}
\newcolumntype{d}[1]{D{.}{.}{#1}}

\newcommand\SetSymbol[1][]{\nonscript\:#1\vert\allowbreak\nonscript\:\mathopen{}}
\providecommand\given{} 
\DeclarePairedDelimiterX\set[1]\{\}{\renewcommand\given{\SetSymbol[\delimsize]}#1}

\DeclareMathOperator{\Var}{Var}

\DeclareMathOperator*{\argmin}{arg\,min}

\graphicspath{{figures/}}

\usepackage{xargs}
\usepackage{xspace}

\usepackage[colorinlistoftodos,prependcaption,textsize=tiny,disable]{todonotes}

\newcommandx{\uvc}[2][1=]{\todo[color=Turquoise!50,#1]{\sf \textbf{\"Umit:} #2}\xspace}

\newcommandx{\mfb}[2][1=]{\todo[color=red!50,#1]{\sf \textbf{Fatih:} #2}\xspace}
\newcommandx{\ks}[2][1=]{\todo[color=red!50,#1]{\sf \textbf{Kaan:} #2}\xspace}

\title{Layer-Neighbor Sampling --- Defusing Neighborhood Explosion in GNNs}

\author{%
  Muhammed Fatih Bal{\i}n\thanks{Part of this work was done during an internship at NVIDIA}
  \thanks{School of Computational Science and Engineering, Georgia Institute of Technology, Atlanta, GA, USA} \\
  \texttt{balin@gatech.edu} \\
  \And
  \"Umit V. \c{C}ataly\"urek\thanks{Amazon Web Services. This publication describes work performed at the Georgia Institute of Technology and is not associated with AWS.}\protect\phantom{\footnotesize 1}\footnotemark[2]\\
  \texttt{umit@gatech.edu} \\
}

\begin{document}

\maketitle

\begin{abstract}
Graph Neural Networks (GNNs) have received significant attention recently, but training them at a large scale remains a challenge.
Mini-batch training coupled with sampling is used to alleviate this challenge.
However, existing approaches either suffer from the neighborhood explosion phenomenon or have poor performance. 
To address these issues, we propose a new sampling algorithm called LAyer-neighBOR sampling (LABOR). 
It is designed to be a  direct replacement for Neighbor Sampling (NS) with the same fanout hyperparameter while sampling up to 7 times fewer vertices, without sacrificing quality.
By design, the variance of the estimator of each vertex matches NS from the point of view of a single vertex.
Moreover, under the same vertex sampling budget constraints, LABOR converges 
faster than existing layer sampling approaches and can use up to 112 times larger batch sizes compared to NS.
\end{abstract}

\section{Introduction}

Graph Neural Networks (GNN)~\cite{Hamilton2017, Kipf2017} have become de facto models
for representation learning on graph structured data. Hence they have started being
deployed in production systems~\cite{Ying2018, Niu2020}. These models iteratively
update the node embeddings by passing messages along the direction of the edges in the
given graph with nonlinearities in between different layers. With $l$ layers, the
computed node embeddings contain information from the $l$-hop neighborhood of the seed vertex.

In the production setting, the GNN models need to be trained on billion-scale graphs~\citep{Ching2015, Ying2018}. 
The training of these models takes hours to days even on distributed systems~\cite{zheng2022distributed, Zheng2022}.
As in general Deep Neural Networks (DNN), it is more
efficient to use mini-batch training~\citep{dnnminibatch} on GNNs,
even though it is a bit trickier in this case. The node embeddings in GNNs
depend recursively on their set of neighbors' embeddings, so when there are $l$ layers,
this dependency spans the $l$-hop neighborhood of the node. Real world graphs usually
have a very small diameter and if $l$ is large, the $l$-hop neighborhood may very well
span the entire graph, also known as
the Neighborhood Explosion Phenomenon (NEP)~\citep{graphsaint-iclr20}.

To solve these issues, researchers proposed sampling a subgraph
of the $l$-hop neighborhood of the nodes in the batch. 
There are mainly three different approaches: Node-based, Layer-based and Subgraph-based methods.
Node-based sampling methods~\citep{Hamilton2017, Chen2018, Liu2020, Zhang2021} sample independently and recursively for each node. 
It was noticed that node-based methods sample subgraphs that
are too shallow, i.e., with a low ratio of number of edges to nodes.
Thus layer-based
sampling methods were proposed~\citep{Chen2018a, Zou2019, Huang2018, Dong2021},
where the sampling for the whole layer is done collectively. 
On the other hand subgraph sampling
methods~\citep{Chiang2019, graphsaint-iclr20, Hu2020a, shaDow, pmlr-v139-fey21a, shi2023lmc} do not use the recursive layer by layer
sampling scheme used in the node- and layer-based sampling methods and instead tend to use the same subgraph for all of the layers.
Some of these sampling methods take the magnitudes of embeddings into account~\citep{Liu2020, Zhang2021, Huang2018}, while others, such as~\cite{Chen2018, Cong2021, pmlr-v139-fey21a, shi2023lmc}, cache the historical embeddings to reduce the variance of the computed approximate embeddings.
There are methods sampling from a vertex cache~\cite{Dong2021} filled with
popular vertices. Most of these approaches are orthogonal to each other and they can be incorporated
into other sampling algorithms.

Node-based sampling methods suffer the most from the NEP but they guarantee a good
approximation for each embedding by ensuring each vertex gets $k$
neighbors which is the only
hyperparameter of the sampling algorithm.
Layer-based sampling methods do not suffer as much from the NEP because number of vertices sampled
is a hyperparameter but they can not guarantee
that each vertex approximation is good enough and also their hyperparameters are hard
to reason with, number of nodes to sample at each
layer depends highly on the graph structure (as the numbers in~\cref{tabl:num_sampled} show).
Subgraph sampling methods usually sample sparser subgraphs compared to their node- and layer-based counterparts.
Hence, in this paper, we focus on the node- and layer-based sampling methods and combine their advantages. 
The major contributions of this work can be listed as follows:
\begin{compactitem}

\item We propose the use of Poisson Sampling for GNNs, taking advantage of its
lower variance and computational efficiency against sampling without replacement.
Applying it to the existing layer sampling method LADIES, we get the superior PLADIES
method outperforming the former by up to 2\% in terms of F1-score.

\item We propose a new sampling algorithm called LABOR, combining advantages of
neighbor and layer sampling approaches using Poisson Sampling. LABOR correlates
the sampling procedures of the given set of seed nodes so that the sampled
vertices from different seeds have a lot of overlap, resulting into a
$7\times$ and $4\times$ reduction in the number of vertices and edges sampled compared
to NS, respectively.
Furthermore, LABOR can sample up to $13\times$ fewer edges compared to LADIES.

\item We experimentally verify our findings, show that our proposed sampling algorithm LABOR outperforms both neighbor sampling and layer sampling approaches.
LABOR can enjoy a batch-size of up to $112\times$ larger than NS while sampling the same number of vertices.
\end{compactitem}

\section{Background}
\label{secl:background}

{\bf Graph Neural Networks: }
Given a directed graph $\mathcal{G} = (V, E)$, where $V$ and $E \subset V \times V$ are vertex and
edge sets respectively, $(t \to s) \in E$ denotes an edge from a source vertex $t \in V$ to a
destination vertex $s \in V$, and $A_{ts}$ denotes the corresponding edge weight if provided. If we have
a batch of seed vertices $S \subset V$, let us define $l$-hop neighborhood
$N^{l}(S)$ for the incoming edges as follows:
\begin{equation}
    N(s) = \{t | (t \to s) \in E \},
    N^{1}(S) = N(S)= \cup_{s \in S} N(s), 
    N^{l}(S) = N(N^{(l-1)}(S))
\end{equation}
Let us also define the degree $d_s$ of vertex $s$ as $d_s = |N(s)|$. To simplify, let's
assume uniform edge weights, $A_{ts} = 1, \forall (t \to s) \in E$. Then, our goal is to estimate the
following for each vertex $s \in S$, where $H^{(l - 1)}_t$ is defined as the embedding of the vertex $t$
at layer $l - 1$, and $W^{(l - 1)}$ is the trainable weight matrix at layer $l - 1$, and $\sigma$ is the nonlinear activation function~\citep{Hamilton2017}:
\begin{equation}
    \label{eq:gnn_model}
    Z^{l}_s = \frac{1}{d_s} \sum_{t \to s} H^{(l - 1)}_t W^{(l - 1)},\; H^{l}_s = \sigma(Z^{l}_s)
\end{equation}

{\bf Exact Stochastic Gradient Descent: }
If we have a node prediction task and $V_t \subseteq V$ is the set of training vertices,
$y_s, s \in V_t$ are the labels of the prediction task, and $\ell$ is the loss function for the prediction
task, then our goal is to minimize the following loss function:
%
    $\frac{1}{|V_t|} \sum_{s \in V_t} \ell(y_s, Z^{l}_s)$.
%
Replacing $V_t$ in the loss function with $S \subset V_t$ for each iteration of gradient descent,
we get stochastic gradient descent for GNNs. However with $l$ layers, the computation dependency is on
$N^{l}(S)$, which reaches large portion of the real world graphs, i.e. $|N^{l}(S)| \approx |V|$, making each iteration costly
both in terms of computation and memory.

{\bf Neighbor Sampling: }
Neighbor sampling approach was proposed by~\cite{Hamilton2017} to approximate $Z^{(l)}_s$ for each $s \in S$ with a subset of
$N^{l}(S)$. Given a fanout hyperparameter $k$, this subset is computed recursively by randomly picking $k$ neighbors for each $s \in S$ from $N(s)$ to form
the next layer $S^{1}$, that is a subset of $N^{1}(S)$. If $d_s \leq k$, then the exact neighborhood $N(s)$ is used. For the next layer, $S^{1}$ is treated as the new set of seed vertices and this procedure is applied recursively.

{\bf Revisiting LADIES, Dependent Layer-based Sampling:}
From now on, we will drop the layer notation and focus on a single layer and also ignore the nonlinearities. Let us define $M_t = H_t W$ as a shorthand notation. Then our goal is to approximate:

\begin{equation}
    H_s = \frac{1}{d_s}\sum_{t \to s} M_t
\end{equation}

If we assign probabilities $\pi_t > 0, \forall t \in N(S)$ and normalize it so that
$\sum_{t \in N(S)} \pi_t = 1$, then use sampling with replacement to sample $T \subset N(S)$ with
$|T| = n$, where $n$ is the number of vertices to sample given as input to the LADIES algorithm and $T$ is a multi-set possibly with multiple copies of the same vertices, and
let $\tilde{d}_s = |T \cap N(s)|$ which is the number of sampled vertices for a given vertex $s$, we
get the following two possible estimators for each vertex $s \in S$:
\begin{subequations}
  \begin{tabularx}{\textwidth}{Xp{2cm}X}
  \begin{equation}
    \label{eq:horvitz}
    H'_s = \frac{1}{nd_s} \sum_{t \in T \cap N(s)} \frac{M_t}{\pi_t}
  \end{equation}
  & &
  \begin{equation}
    \label{eq:hajek}
    H''_s = \frac{\sum_{t \in T \cap N(s)} \frac{M_t}{\pi_t}}{\sum_{t \in T \cap N(s)} \frac{1}{\pi_t}}
  \end{equation}
  \end{tabularx}
\end{subequations}

Note that $H'_s$ in~(\ref{eq:horvitz}) is the Thompson-Horvitz estimator and the $H''_s$ 
in~(\ref{eq:hajek}) is the Hajek estimator. 
For a comparison between the two and how to get an even better estimator by combining them, see~\cite{Khan2021}. 
The formulation in the LADIES paper uses $H'_s$, but it proposes to row-normalize the sampled adjacency matrix,
meaning they use $H''_s$ in their implementation.
However, analyzing the variance of the Thompson-Horvitz estimator is simpler and its variance serves as an upper bound for the variance of the Hajek estimator when $|M_t|$ and $\pi_t$ are uncorrelated~\cite{Khan2021, Dorfman1997}, which we assume to be true in our case. Note that the variance analysis is simplified to be element-wise for all vectors involved.
\begin{equation}
    \label{eq:variance_ht}
    \Var(H''_s) \leq \Var(H'_s) = \frac{1}{\tilde{d}_s d_s^2} \sum_{t \to s} \pi_t \sum_{t' \to s} \frac{\Var(M_{t'})}{\pi_{t'}}
\end{equation}

Since we do not have access to the computed embeddings and to simplify the analysis, 
we assume that $\Var(M_t) = 1$ from now on. One can see that $\Var(H'_s)$ is minimized when 
$\pi_t = p, \forall t \to s$ under the constraint $\sum_{t \to s} \pi_t \leq p d_s$ for some constant $p \in [0, 1]$, hence any deviation from uniformity increases the variance. 
The variance is also smaller the larger $\tilde{d}_s$ is. However, in theory and in practice, there is 
no guarantee that each vertex $s \in S$ will get
any neighbors in $T$, not to mention equal numbers of neighbors. Some vertices will have pretty good 
estimators with thousands of samples and very low variances,
while others might not even get a single neighbor sampled. For this reason, we designed LABOR so 
that every vertex in $S$ will sample enough neighbors in expectation.

While LADIES is optimal from an approximate matrix multiplication perspective~\cite{Chen2022},
it is far from optimal in the case of nonlinearities and multiple layers.
Even if there is a single layer, then the used loss functions are nonlinear. Moreover,
the existence of nonlinearities in-between layers and the fact that there are multiple layers exacerbates this issue and
necessitates that each vertex gets a good enough estimator with low enough variance.
Also, LADIES gives a formulation using sampling with replacement instead of 
without replacement and that is sub-optimal from the variance perspective while its implementation
uses sampling without replacement without taking care of the bias created thereby. In the next section, we 
will show how all of these problems are addressed by our newly proposed Poisson sampling framework and LABOR sampling.

\section{Proposed Layer Sampling Methods}

Node-based sampling methods suffer
from sampling too shallow subgraphs leading to NEP in just a few hops (e.g., see~\cref{tabl:num_sampled}). Layer sampling methods
\cite{Zou2019}
attempt to fix this by sampling a fixed number of vertices in each layer, however they can not ensure that the
estimators for the vertices are of high quality, and it is hard to reason how to choose the number 
of vertices to sample in each layer.
LADIES~\cite{Zou2019} proposes using the same
number for each layer while papers evaluating it found it is better to sample an increasing 
number of vertices in each layer~\cite{Liu2020, Chen2022}.
There is no systematic way to choose how many vertices to sample in each layer for the LADIES method, and since each graph has different density and connectivity structure, this choice highly depends on the graph in question. 
Therefore, due to its simplicity and high quality results, Neighbor Sampling currently seems to be the most popular
sampling approach and there exists high quality implementations on both CPUs and 
GPUs in the popular GNN frameworks~\cite{wang2019deep, fey2019fast}.

We propose a 
new approach that combines the advantages of layer and neighbor sampling approaches
using a vertex-centric variance based framework,
reducing the number of sampled vertices drastically while ensuring the training quality does not 
suffer and matches the quality of neighbor sampling.
Another advantage of our method is that the user only needs to choose the batch size and the fanout 
hyperparameters as in the
Neighbor Sampling approach, the algorithm itself then samples the minimum number of vertices in the 
later layers in an unbiased way while ensuring each vertex gets enough neighbors and a good 
approximation.

We achieve all the previously mentioned good properties with the help of Poisson Sampling.
So, next section will demonstrate applying Poisson Sampling to Layer Sampling, then we will show
how the advantages of Layer and Neighbor Sampling methods can be combined into LABOR while getting rid of their cons altogether.

\subsection{Poisson Layer Sampling (PLADIES)}
\label{subsecl:pladies}

In layer sampling, the main idea can be summarized
as individual vertices making correlated decisions while sampling their neighbors, because in the 
end if a vertex $t$ is sampled, all edges into the seed vertices $S$, i.e., $t
\to s$, $s \in S$, are added to the sampled subgraph.
This can be interpreted as vertices in $S$ making a collective decision on
whether to sample $t$, or not.

The other thing to keep in mind is that, the existing layer sampling methods use sampling with 
replacement when doing importance sampling with unequal probabilities, because it is nontrivial to
compute the inclusion probabilities in the without replacement case. The Hajek estimator in 
the without replacement case with equal probabilities becomes:
\begin{equation}
    H''_s = \frac{\sum_{t \in T \cap N(s)} \frac{M_t}{\bar{\pi}_t}}{\sum_{t \in T \cap N(s)} \frac{1}{\bar{\pi}_t}} = \frac{\sum_{t \in T \cap N(s)} M_t |N(S)|}{\sum_{t \in T \cap N(s)} |N(S)|} = \frac{1}{\tilde{d}_s}\sum_{t \in T \cap N(s)} M_t
\end{equation}
and it has the variance:
\begin{equation}
    \label{eq:neighbor_sampling_variance}
    \Var(H''_s) = \frac{d_s - \tilde{d}_s}{d_s - 1} \frac{1}{\tilde{d}_s}
\end{equation}
%
Let us show how one can do layer sampling using Poisson sampling (PLADIES). 
Given probabilities $\pi_t \in [0, 1], \forall t \in N(S)$ so that $\sum_{t \in N(S)} \pi_t = n$, we 
include $t \in N(S)$ in our sample $T$ with probability $\pi_t$ by flipping a coin for it, i.e., we sample $r_t \sim U(0, 1)$ and include $t \in T$ if $r_t \leq \pi_t$.
In the end, $E[|T|] = n$ and we can still use the Hajek estimator $H''_s$ or the Horvitz 
Thomson estimator $H'_s$ to estimate $H_s$. Doing layer sampling this way is unbiased by 
construction and achieves the same goal in linear time in contrast to the quadratic time debiasing approach 
explained in~\cite{Chen2022}. The variance then approximately becomes~\citep{Williams1998},
see ~\cref{subsecl:poisson_sampling_variance_derivation} for a derivation:
\begin{equation}
    \label{eq:poisson_sampling_variance}
    \Var(H''_s) \leq \Var(H'_s) = \frac{1}{d_s^2} \sum_{t \to s} \frac{1}{\pi_t} - \frac{1}{d_s}
\end{equation}
One can notice that the minus term $\frac{1}{d_s}$ enables the variance to 
converge to $0$, if all $\pi_t = 1$ and we get the exact result.
However, in the sampling with replacement case, the variance goes to $0$ only as the sample size goes 
to infinity.

\subsection{LABOR: \texorpdfstring{\underline{La}}{La}yer Neigh\texorpdfstring{\underline{bor}}{bor} Sampling}
\label{subsecl:labor}

The design philosophy of LABOR Sampling is to create a direct alternative to Neighbor Sampling while 
incorporating the advantages of layer sampling. 
Mimicking Layer Sampling with Poisson Sampling in~\cref{subsecl:pladies} still has the disadvantage that 
$\tilde{d}_s$ varies wildly for different $s$. To overcome this and mimic Neighbor Sampling where 
$E[\tilde{d}_s] = \min(d_s, k)$, where $k$ is a given fanout hyperparameter, we proceed as follows: for given 
$\pi_t \geq 0, \forall t \in N(S)$ denoting unnormalized probabilities, for a given $s$, let us define $c_s$ as the 
quantity satisfying the following equality if $k < d_s$, otherwise $c_s = \max_{t \to s} \frac{1}{\pi_t}$:
\begin{equation}
    \label{eq:variance_poisson}
    \frac{1}{d_s^2}\sum_{t \to s} \frac{1}{\min(1, c_s \pi_t)} - \frac{1}{d_s} = \frac{1}{k} - \frac{1}{d_s}
\end{equation}

Note that $\frac{1}{k} - \frac{1}{d_s}$ is the variance when
$\pi_t = \frac{k}{d_s}, \forall t \in N(s)$ so that $E[\tilde{d}_s] = k$. Also note that:
\begin{equation}
    \frac{d_s}{d_s - 1}(\frac{1}{k} - \frac{1}{d_s}) - \frac{d_s - k}{d_s - 1} \frac{1}{k} = \frac{d_s - k}{k (d_s - 1)} - \frac{d_s - k}{d_s - 1} \frac{1}{k} = 0
\end{equation}

meaning that the variance target we set through~(\ref{eq:variance_poisson}) is equal to Neighbor Sampling's
variance in~(\ref{eq:neighbor_sampling_variance}) after calibrating with $\frac{d_s}{d_s - 1}$~\cite{Ohlsson1998}
and it will result 
in $E[\tilde{d}_s] \geq k$ with strict equality in the uniform probability case.
Then each vertex $s \in S$ samples $t \to s$ with probability $c_s \pi_t$. To keep the collective 
decision making, we sample $r_t \sim U(0, 1), \forall t \in N(S)$
and vertex $s$ samples vertex $t$ if and only if $r_t \leq c_s \pi_t$. Note that if we use a uniform random variable for each edge $r_{ts}$ instead of each vertex $r_t$, and if $\pi$ is uniformly initialized, then we get
the same behavior as Neighbor Sampling.

\subsubsection{Importance Sampling}

Given the sampling procedure above, one wonders how different choices of $\pi
\geq 0$ will affect $|T|$, the total number of unique vertices sampled. We can
compute as follows: 
\begin{equation}
    E[|T|] = \sum_{t \in N(S)} \mathbb{P}(t \in T) = \sum_{t \in N(S)} \min(1, \pi_t \max_{t \to s} c_s)
\end{equation}

In particular, we need to find $\pi^* \geq 0$ minimizing $E[|T|]$:
\begin{equation}
    \label{eq:objective_function}
     \pi^* = \argmin_{\pi \geq 0} \sum_{t \in N(S)} \min(1, \pi_t \max_{t \to s} c_s)
\end{equation}
Note that for any given $\pi \geq 0$, $E[|T|]$ is the same for any vector
multiple $x \pi, x \in \mathbb{R}^+$, meaning that the objective function is
homogenous of degree $0$.

\subsubsection{Computing \texorpdfstring{$c$}{c} and \texorpdfstring{$\pi^*$}{pi}}
\label{subsecl:optimizing}

$c_s$ was defined to be the scalar satisfying the following equality involving
the variance of the estimator of $H_s$:
\begin{equation}
    \frac{1}{d_s^2}\sum_{t \to s} \frac{1}{\min(1, c_s \pi_t)} - \frac{1}{d_s} = \frac{1}{k} - \frac{1}{d_s}
\end{equation}
If we rearrange the terms, we get:
\begin{equation}
    \sum_{t \to s} \frac{1}{\min(1, c_s \pi_t)} = \frac{d_s^2}{k}
\end{equation}
One can see that the left hand side of the equality is monotonically decreasing with respect to $c_s \geq 0$.
Thus one can use binary search to find the $c_s$ satisfying the 
above equality to any precision needed. But we opt to use the following iterative algorithm to compute it:
\begin{gather}
    v_s^{(0)} = 0, c_s^{(0)} = \frac{k}{d_s^2} \sum_{t \to s} \frac{1}{\pi_t} \\
    c_s^{(i + 1)} = \frac{c_s^{(i)}}{\frac{d_s^2}{k} - v_s^{(i)}}\Big( -v_s^{(i)} + \sum_{t \to s} \frac{1}{\min(1, c_s^{(i)} \pi_t)} \Big) \\
    v_s^{(i + 1)} = \sum_{t \to s} \mathbbm{1}[c_s^{(i + 1)} \pi_t \geq 1]
\end{gather}

This iterative algorithm converges in at most $d_s$ steps and the convergence is exact and monotonic from below.
One can also implement it in linear time $\mathcal{O}(d_s)$ if $\set{\pi_t \given t \to s}$ is sorted and
making use of precomputed prefix sum arrays. Note that $c = c(\pi)$, meaning that $c$ is a function of the given probability vector $\pi$.
To compute $\pi^*$, we use a similar fixed point iteration as follows:
\begin{equation}
    \pi^{(0)} = 1, \forall t \in N(S): \pi^{(i + 1)}_t = \pi_t^{(i)} \max_{t \to s} c_s(\pi^{(i)})
    \label{eq:pi_fixed_point_iteration}
\end{equation}
Thus, we alternate between computing $c = c(\pi)$, meaning $c$ is computed with the current $\pi$, and updating 
the $\pi$ with the computed $c$ values. Each step of this iteration is guaranteed to lower the objective 
function value in~(\ref{eq:objective_function}) until convergence to a fixed point, see
the ~\cref{subsecl:fixed_point_iterations_proof}. Modified formulation for a given nonuniform weight matrix $A_{ts}$ is explained in the~\cref{subsecl:weighted_extension}.~\cref{alg:labor} in the Appendix summarizes LABOR.

\begin{table*}[ht]
    \centering
    \begin{small}
    \caption{Properties of the datasets used in experiments: numbers of vertices
    ($|V|$), edges ($|E|$), avg. degree ($\frac{|E|}{|V|}$), number of features, sampling budget used, training, validation and test vertex split.} 
    \label{tabl:dataset}
    \begin{tabular}{c | r | r | r | r | r | r}
    \toprule
    \textbf{Dataset} & \textbf{$\bm{|V|}$} & \textbf{$\bm{|E|}$} & \textbf{$\bm{\frac{|E|}{|V|}}$} & \textbf{\# feats.} & \textbf{$\bm{|V^3|}$ budget} & \textbf{train - val - test (\%)} \\
    \midrule
    reddit & 233K & 115M & 493.56 & 602 & 60k & 66 - 10 - 24\\
    products & 2.45M & 61.9M & 25.26 & 100 & 400k & 8 - 2 - 90 \\
    yelp & 717K & 14.0M & 19.52 & 300 & 200k & 75 - 10 - 15 \\
    flickr & 89.2K & 900K & 10.09 & 500 & 70k & 50 - 25 - 25 \\
    \bottomrule
    \end{tabular}
    \end{small}
\end{table*}

\section{Experiments}
\label{secl:experiments}

In this section, we empirically evaluate the performance of each method in the node-prediction 
setting on the following datasets: reddit~\citep{Hamilton2017},
products~\citep{Hu2020}, yelp, flickr~\citep{graphsaint-iclr20}. Details about
these datasets are given in~\cref{tabl:dataset}.
We compare our proposed LABOR variants LABOR-0, LABOR-1 and LABOR-*,
where $0, 1, *$ stand for the number of fixed point iterations applied to
optimize~(\ref{eq:objective_function}), respectively, together with PLADIES (see~\cref{subsecl:pladies}),
against the baseline sampling methods: Neighbor Sampling (NS) and LADIES.
We do not include Fast-GCN in our comparisons as it is super-seeded by the LADIES paper.
The works of \cite{Liu2020, Zhang2021, Huang2018, Cong2021, Dong2021} are not included in the
comparisons because they either take into account additional information such as historical embeddings
or their magnitudes or they have an additional sampling structure such as a vertex cache to sample from.
Also the techniques in these papers are mostly orthogonal to the sampling
problem and algorithms discussed in this paper, hence our proposed LABOR and PLADIES methods can
be used together with them. However, we leave this investigation to future work.

We evaluate all the methods on the GCN model in~(\ref{eq:gnn_model}) with 3
layers, with 256 hidden dimension and residual skip connections enabled.
We use the Adam optimizer~\citep{Kingma2014} with $0.001$ learning rate.
We implemented LABOR variants and PLADIES in DGL~\citep{wang2019deep},
and carried out our experiments using the DGL with the Pytorch
backend~\citep{Paszke2019PyTorchAI}.
\footnote{Our implementation is available in DGL starting with version 1.0 as \texttt{dgl.dataloading.LaborSampler} and \texttt{dgl.sampling.sample\_labors}.}
Experiments were repeated 100 times and averages are presented.

\subsection{Baseline Comparison of Vertex/Edge Efficiency}
\label{subsecl:NS_comp}

\begin{figure*}[!h]
    \centering
    \includegraphics[width=\linewidth]{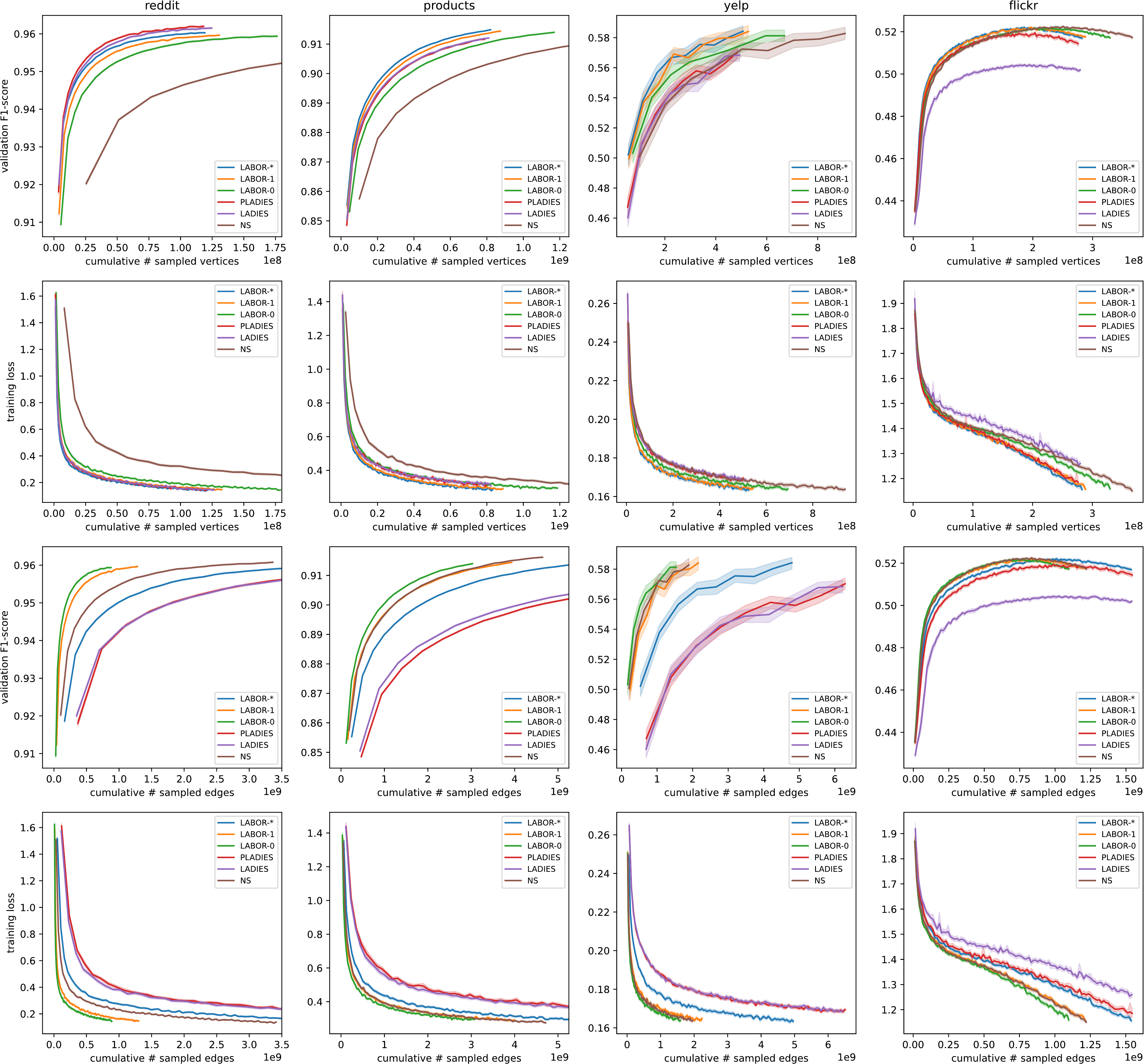}
    \caption{The validation F1-score and training loss curves on different
    datasets with same batch size. The x-axis stands for cumulative number of
    vertices and edges used during training, respectively. The soft edges
    represent the confidence interval. Number of sampled vertices and edges can
    be found in~\cref{tabl:num_sampled}.}
    \label{figl:same_loss_cum_both}
\end{figure*}

\begin{table*}[ht]
    \centering
    \begin{small}
\caption{Average number of vertices and edges sampled in different layers (All
the numbers are in thousands, lower is better). Last two columns show iterations
(mini-batches) per second (it/s) and test F1-score, for both, higher is better. The hyperparameters of
LADIES and PLADIES were picked to roughly match the number of vertices sampled
by the LABOR-*  to get a fair comparison. The convergence curves can be found
in~\cref{figl:same_loss_cum_both}.  The timing information was
measured on an NVIDIA T4 GPU. {\color{OliveGreen}{\textbf{Green}}} stands for best,
{\color{Maroon}{\textbf{red}}} stands for worst results, with a $5\%$ cutoff.}
\label{tabl:num_sampled}
\begin{tabular}{c | c | r r r r r r r | r | r}
\toprule
\textbf{Dataset} & \textbf{Algo.} & $\bm{|V^3|}$ & $\bm{|E^2|}$ & $\bm{|V^2|}$ & $\bm{|E^1|}$ & $\bm{|V^1|}$ & $\bm{|E^0|}$ & $\bm{|V^0|}$ & \textbf{it/s} & \textbf{test F1-score} \\
\midrule
 & PLADIES & \color{OliveGreen}{\textbf{24}} & \color{Maroon}{\textbf{2390}} & 14.1 & 927 & 6.0 & 33.2 & 1 & \color{Maroon}{\textbf{1.7}} & 96.21 $\pm$ 0.06 \\
 & LADIES & \color{OliveGreen}{\textbf{25}} & \color{Maroon}{\textbf{2270}} & 14.5 & 852 & 6.0 & 32.5 & 1 & \color{Maroon}{\textbf{1.8}} & 96.20 $\pm$ 0.05 \\
reddit & LABOR-* & \color{OliveGreen}{\textbf{24}} & 1070 & 13.7 & 435 & 6.0 & 26.9 & 1 & 4.1 & 96.23 $\pm$ 0.05 \\
 & LABOR-1 & 27 & 261 & 14.4 & 116 & 6.1 & 16.7 & 1 & 24.8 & 96.23 $\pm$ 0.06 \\
 & LABOR-0 & 36 & \color{OliveGreen}{\textbf{177}} & 17.8 & 67 & 6.8 & 9.6 & 1 & \color{OliveGreen}{\textbf{37.6}} & 96.25 $\pm$ 0.05 \\
 & NS & \color{Maroon}{\textbf{167}} & 682 & 68.3 & 100 & 10.1 & 9.7 & 1 & 14.2 & 96.24 $\pm$ 0.05 \\
\midrule
 & PLADIES & \color{OliveGreen}{\textbf{160}} & \color{Maroon}{\textbf{2380}} & 51.2 & 293 & 9.7 & 11.7 & 1 & \color{Maroon}{\textbf{4.1}} & 78.44 $\pm$ 0.24 \\
 & LADIES & \color{OliveGreen}{\textbf{165}} & \color{Maroon}{\textbf{2230}} & 51.8 & 270 & 9.7 & 11.5 & 1 & \color{Maroon}{\textbf{4.2}} & 78.59 $\pm$ 0.22 \\
products & LABOR-* & \color{OliveGreen}{\textbf{166}} & 1250 & 51.8 & 167 & 9.8 & 10.6 & 1 & 6.2 & 78.59 $\pm$ 0.34 \\
 & LABOR-1 & 178 & 799 & 53.4 & 136 & 9.8 & 10.5 & 1 & 21.3 & 78.47 $\pm$ 0.26 \\
 & LABOR-0 & 237 & \color{OliveGreen}{\textbf{615}} & 62.4 & 100 & 10.1 & 9.9 & 1 & \color{OliveGreen}{\textbf{32.5}} & 78.76 $\pm$ 0.26 \\
 & NS & \color{Maroon}{\textbf{513}} & 944 & 95.4 & 106 & 10.6 & 9.9 & 1 & 24.6 & 78.48 $\pm$ 0.29 \\
\midrule
 & PLADIES & \color{OliveGreen}{\textbf{100}} & \color{Maroon}{\textbf{1300}} & 29.5 & 183 & 6.2 & 6.9 & 1 & \color{Maroon}{\textbf{5.1}} & 61.55 $\pm$ 0.87 \\
 & LADIES & \color{OliveGreen}{\textbf{102}} & \color{Maroon}{\textbf{1280}} & 29.7 & 182 & 6.2 & 6.9 & 1 & \color{Maroon}{\textbf{5.3}} & 61.89 $\pm$ 0.66 \\
yelp & LABOR-* & \color{OliveGreen}{\textbf{105}} & 991 & 30.7 & 158 & 6.1 & 6.8 & 1 & 13.3 & 61.57 $\pm$ 0.67 \\
 & LABOR-1 & 109 & 447 & 31.0 & 96 & 6.2 & 6.8 & 1 & \color{OliveGreen}{\textbf{27.3}} & 61.71 $\pm$ 0.70 \\
 & LABOR-0 & 138 & \color{OliveGreen}{\textbf{318}} & 35.1 & 54 & 6.2 & 6.3 & 1 & \color{OliveGreen}{\textbf{27.2}} & 61.55 $\pm$ 0.85 \\
 & NS & \color{Maroon}{\textbf{188}} & 392 & 42.5 & 55 & 6.3 & 6.3 & 1 & 23.0 & 61.50 $\pm$ 0.66 \\
\midrule
 & PLADIES & \color{OliveGreen}{\textbf{55}} & \color{Maroon}{\textbf{309}} & 24.9 & 85 & 6.2 & 6.9 & 1 & \color{Maroon}{\textbf{10.2}} & 51.52 $\pm$ 0.26 \\
 & LADIES & \color{OliveGreen}{\textbf{56}} & \color{Maroon}{\textbf{308}} & 25.1 & 85 & 6.2 & 6.9 & 1 & \color{Maroon}{\textbf{10.5}} & \color{Maroon}{\textbf{50.79 $\pm$ 0.29}} \\
flickr & LABOR-* & \color{OliveGreen}{\textbf{57}} & \color{Maroon}{\textbf{308}} & 25.6 & 85 & 6.3 & 6.9 & 1 & 20.3 & 51.67 $\pm$ 0.27 \\
 & LABOR-1 & 58 & 242 & 25.9 & 73 & 6.3 & 6.9 & 1 & \color{OliveGreen}{\textbf{32.7}} & 51.66 $\pm$ 0.24 \\
 & LABOR-0 & 66 & \color{OliveGreen}{\textbf{219}} & 29.1 & 52 & 6.4 & 6.7 & 1 & \color{OliveGreen}{\textbf{33.3}} & 51.65 $\pm$ 0.26 \\
 & NS & \color{Maroon}{\textbf{73}} & 244 & 32.8 & 52 & 6.4 & 6.7 & 1 & \color{OliveGreen}{\textbf{31.7}} & 51.70 $\pm$ 0.23 \\
\bottomrule
\end{tabular}
\end{small}
\end{table*}

\paragraph{LABOR comparison against NS:}
In this experiment, we set the batch size to 1,000
and the fanout $k = 10$ for LABOR and NS methods to
see the difference in the sizes of the sampled subgraphs and also how vertex/edge efficient they are.
In~\cref{figl:same_loss_cum_both}, we can see that LABOR variants outperform NS
in both vertex and edge efficiency metrics.
\cref{tabl:num_sampled} shows the difference of the sampled subgraph sizes in each layer. One can see that
on reddit, LABOR-* samples $6.9 \times$ fewer vertices in the 3rd layer leading much faster 
convergence in terms of cumulative number of sampled vertices. On
the flickr dataset however, LABOR-* samples only $1.3 \times$ fewer vertices. The amount of
difference depends on two factors. The first is the amount of overlap of neighbors among the vertices in $S$.
If the neighbors of vertices in $S$ did not overlap at all, then one obviously can not do better than NS.
The second is the average degree of the graph. With a fanout of $10$, both NS and LABOR has to
copy the whole neighborhood of a vertex $s$ with degree $d_s \leq 10$. Thus for such graphs, it is expected 
that the difference will be insignificant.
The average degree of the flickr
is $10.09$ (\cref{tabl:dataset}), and thus there is only a small difference between LABOR and NS.

As seen in~\cref{figl:same_loss_cum_both} and~\cref{tabl:num_sampled},
LABOR-0 reduces both the number of vertices and edges sampled. On the other hand, 
when importance sampling is enabled, the number of vertices sampled goes down while number of edges
sampled goes up. This is because when importance sampling is used, inclusion probabilities become
nonuniform and it takes more edges per seed vertex to get a good approximation
(see~\cref{eq:variance_poisson}). Thus, LABOR-0 leads the pack when it comes to edge efficiency.

\paragraph{Layer sampling algorithm comparison:}
The hyperparameters of LADIES and PLADIES were picked to match LABOR-* so that all methods have the same 
vertex sampling budget in each layer (see~\cref{tabl:num_sampled}).
\cref{figl:same_loss_cum_both} shows that, in terms of convergence, LADIES and PLADIES perform almost the 
same as each other, on all but the flickr dataset, in which case there is a big difference between the two in favor of 
PLADIES. We also see that LABOR variants outperform LADIES variants on products, yelp and flickr in 
terms of vertex efficiency with the exception of reddit, where the best performing algorithm is 
PLADIES proposed by us. When it comes to edge efficiency however, we see that LADIES variants
are simply inefficient and all LABOR variants outperform LADIES variants on across all the datasets
by up to $13\times$.

\paragraph{Comparing LABOR variants:}
Looking at~\cref{tabl:num_sampled}, we can see that LABOR-0 has the best runtime performance across all
datasets. 
This is due to 1)~sampling fewer edges, 2)~ not having the fixed point iteration overhead, compared to the other LABOR variants. By design, all LABOR variants should
have the same convergence curves, as seen in~\cref{subsecl:same_convergence} in~\cref{figl:same_loss}. Then, the decision of which variant to
use depends on one factor: feature access speed. If vertex features were stored on a
slow storage medium (such as, on host memory accessed over PCI-E), then minimizing number of sampled vertices would become the highest priority, in which
case, one should pick LABOR-*. Depending on the relative vertex feature access performance and the performance
of the training processor, one can choose to use LABOR-$j$, the faster feature access, the lower the $j$.

\begin{figure*}[!h]
    \centering
    \includegraphics[width=\linewidth]{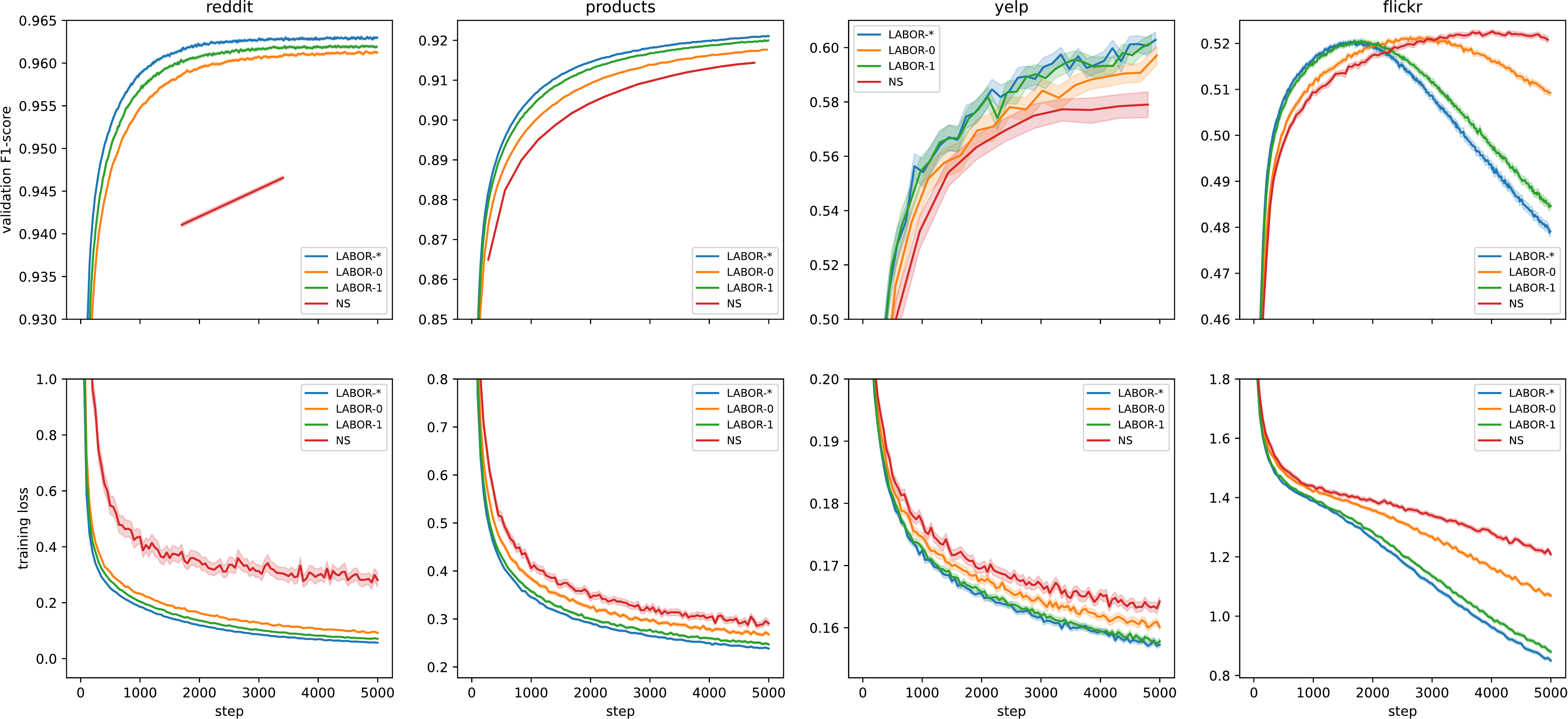}
    \vspace*{-2ex}
    \caption{Validation F1-score and the training loss curves to evaluate vertex sampling efficiency under the same sampling budget. The x-axis stands for the number of training iterations. The batch size is chosen so that the \# sampled vertices matches the vertex budget for each dataset and method and it is given in~\cref{tabl:batch_size}.}
    \label{figl:same_budget}
\end{figure*}

\subsection{LABOR vs NS under vertex sampling budgets}
\label{subsecl:vertex_efficiency}

In this experiment, we set a limit on the number of sampled vertices and modify the 
batch size to match the given vertex budget. The budgets used 
were picked around the same magnitude with numbers in the~\cref{tabl:num_sampled} in the $|V_3|$ column 
and can be found in~\cref{tabl:dataset}.
\cref{figl:same_budget} displays the result of this experiment. \cref{tabl:batch_size} shows that the more vertex efficient the sampling method is, the larger batch size it can use during 
training. Number of sampled vertices is not a function of the batch size for the LADIES algorithm so we do 
not include it in this comparison. All of the experiments were repeated 100 times and their averages were 
plotted, that is why our convergence plots are smooth and differences are clear.
The most striking result in this experiment is that there can be up to $112\times$ difference in batch sizes of LABOR-* and NS 
algorithms on the reddit dataset, which translates into faster convergence as the training loss and validation F1-score curves in~\cref{figl:same_budget} show.

\begin{table}
\parbox{.45\linewidth}{
\centering
\begin{small}
\caption{The batch sizes used in~\cref{figl:same_budget}. These were chosen such that in expectation, each method
samples with the same budget given in~\cref{tabl:dataset}. Having a larger batch-size speeds up convergence.}
\label{tabl:batch_size}
\begin{tabular}{c | r | r | r | r}
\toprule
\textbf{Dataset} & \textbf{LAB-*} & \textbf{LAB-1} & \textbf{LAB-0} & \textbf{NS} \\
\midrule
reddit & \color{OliveGreen}{\textbf{10100}} & 8250 & 3850 & \color{Maroon}{\textbf{90}} \\
products & \color{OliveGreen}{\textbf{7750}} & 6000 & 2750 & \color{Maroon}{\textbf{700}} \\
yelp & \color{OliveGreen}{\textbf{3700}} & 3300 & 1950 & \color{Maroon}{\textbf{1120}} \\
flickr & \color{OliveGreen}{\textbf{3600}} & 3100 & 1400 & \color{Maroon}{\textbf{800}} \\
\bottomrule
\end{tabular}
\end{small}
}
\hfill
\parbox{.5\linewidth}{
\centering
\begin{small}
\caption{Number of vertices (in thousands) in 3rd layer w.r.t \# fixed point iterations (its). * denotes applying the fixed point iterations until convergence, i.e., LABOR-*, 1 its stands for LABOR-1 etc.}
\label{tabl:imp_num_sampled}
\begin{tabular}{c | r | r r r r r}
\toprule
\textbf{Dataset} & \multicolumn{1}{c|}{\textbf{NS}} & \multicolumn{1}{c}{\textbf{0}} & \multicolumn{1}{c}{\textbf{1}} & \multicolumn{1}{c}{\textbf{2}} & \multicolumn{1}{c}{\textbf{3}} & \multicolumn{1}{c}{\textbf{*}} \\
\midrule
reddit & 167 & 36 & 27 & 25 & 25 & 24 \\
products & 513 & 237 & 178 & 170 & 169 & 166 \\
yelp & 188 & 138 & 109 & 106 & 105 & 105 \\
flickr & 73 & 66 & 58 & 57 & 57 & 56 \\
\bottomrule
\end{tabular}
\end{small}
}
\end{table}

\subsection{Importance Sampling, Fixed Point Iterations}
\label{subsecl:imp_sampling}

In this section, we look at the convergence behavior of the fixed point iterations described in~\cref{subsecl:optimizing}.
\cref{tabl:imp_num_sampled} shows the number of sampled vertices in the 
last layer with respect to the number of fixed point iterations applied. In this table, the * stands for 
applying the fixed point iterations until convergence, and convergence occurs in at most 15 iterations in 
practice before the relative change in the objective function is less than $10^{-4}$. One can see that most of 
the reduction in the objective function~(\ref{eq:objective_function}) occurs after the first iteration, and the 
remaining iterations have diminishing returns. Full convergence can save 
from $14\%$ - $33\%$ depending on the dataset. The monotonically decreasing numbers provide empirical evidence for the presented 
proof in the~\cref{subsecl:fixed_point_iterations_proof}.
%


\section{Conclusions}

In this paper, we introduced LABOR sampling, a novel way to combine layer and neighbor sampling approaches 
using a vertex-variance centric framework. We then transform the sampling problem into an optimization 
problem where the constraint is to match neighbor sampling variance for each vertex while sampling the fewest 
number of vertices. We show how to minimize this new objective function via fixed-point iterations. On
datasets with dense graphs like Reddit, we show that our approach can sample a subgraph with $7 \times$ fewer vertices without 
degrading the batch quality. We also show that compared to LADIES, LABOR converges faster with same sampling budget.

\section*{Acknowledgements}

We would like to thank Dominique LaSalle and Kaan Sancak for their feedback on the manuscript, and also
Murat Guney for their helpful discussions and enabling this project during the author's time at NVIDIA.
This work was partially supported by
the NSF grant CCF-1919021.

\bibliographystyle{plainnat}
\bibliography{main}

\newpage
\appendix
\section{Appendix}
\label{secl:appendix}

\begin{multicols}{2}

\subsection{Derivation of Poisson Sampling variance}
\label{subsecl:poisson_sampling_variance_derivation}

Following is a derivation of~\cref{eq:poisson_sampling_variance} for the Horvitz-Thomson estimator. Note that we
assume $\Var(M_t) = 1, \forall t$. Also, $\Var(\mathbbm{1}[r_t \leq \pi_t])$ is the variance of the bernoulli
trial with probability $\pi_t$ and is equal to $\pi_t (1 - \pi_t)$.

\begin{equation}
\begin{aligned}
    H'_s &= \frac{1}{d_s} \sum_{t \to s} \frac{M_t}{\pi_t} \mathbbm{1}[r_t \leq \pi_t] \\
    \Var(H'_s) &= \Var\Big( \frac{1}{d_s} \sum_{t \to s} \frac{M_t}{\pi_t} \mathbbm{1}[r_t \leq \pi_t] \Big) \\ 
    &= \frac{1}{d_s^2} \sum_{t \to s} \frac{\Var(M_t)}{\pi_t^2} \Var(\mathbbm{1}[r_t \leq \pi_t]) \\
    &= \frac{1}{d_s^2} \sum_{t \to s} \frac{\Var(M_t)}{\pi_t^2} \pi_t (1 - \pi_t) \\
    &= \frac{1}{d_s^2} \sum_{t \to s} \frac{\Var(M_t)}{\pi_t} (1 - \pi_t) \\
    &= \frac{1}{d_s^2} \sum_{t \to s} \frac{1}{\pi_t} (1 - \pi_t) \\
    &= \frac{1}{d_s^2} \sum_{t \to s} (\frac{1}{\pi_t} - 1) \\
    &= \frac{1}{d_s^2} \sum_{t \to s} \frac{1}{\pi_t} - \frac{1}{d_s}
\end{aligned}
\end{equation}

\begin{algorithm}[H]
   \caption{LABOR-i for uniform edge weights}
   \label{alg:labor}
\begin{algorithmic}
   \STATE {\bfseries Input:} seed vertices $S$, \# iterations $i$, fanout $k$
   \STATE {\bfseries Output:} sampled edges $E'$, sampled weights $A'$
   \STATE $T \gets \{ t \mid t \in N(S) \}$
   \STATE $\pi_t \gets 1, \forall t \in T$
   \WHILE{$i > 0$}
     \FORALL{$s \in S$}
        \STATE Solve (14) for $c_s$
     \ENDFOR
     \FORALL{$t \in T$}
        \STATE $\pi_t \gets \pi_t \max_{t \to s} c_s$
     \ENDFOR
     \STATE $i \gets i - 1$
   \ENDWHILE{}
   \STATE $r_t \sim U(0, 1), \forall t \in T$
   \STATE $E' \gets \text{[ ]}$
   \STATE $A' \gets \text{[ ]}$
   \FORALL{$s \in S$}
     \STATE $w \gets 0$
     \FORALL{$t \in N(s)$}
     \IF{$r_t \leq c_s \pi_t$}
        \STATE $E'$.append($t \to s$)
        \STATE $w \gets w + \frac{1}{\pi_t}$
     \ENDIF
     \ENDFOR
     \FORALL{$t \in N(s)$}
     \IF{$r_t \leq c_s \pi_t$}
        \STATE $A'$.append($\frac{1}{\pi_t w}$)
     \ENDIF
     \ENDFOR
   \ENDFOR
\end{algorithmic}
\end{algorithm}

\end{multicols}

\subsection{Choosing how many neighbors to sample}
\label{subsecl:choosing_how_many}

The variance of Poisson Sampling when $\pi_t = \frac{k}{d_s}$ is $\frac{1}{k} - \frac{1}{d_s}$.
One might question why we are trying to match the variance of Neighbor Sampling and choose
to use a fixed fanout for all the seed vertices. In the uniform probability case, if we have
already sampled some set of edges for all vertices in $S$, and want to sample one more edge,
the question becomes which vertex in $S$ should we sample the new edge for? Our answer to this
question is the vertex $s$, whose variance would improve the most. If currently vertex $s$ has
$\tilde{d}_s$ edges sampled, then sampling one more edge for it would improve its variance
from $\frac{1}{\tilde{d}_s} - \frac{1}{d_s}$ to $\frac{1}{1 + \tilde{d}_s} - \frac{1}{d_s}$. 
Since the derivative of the variance with respect to $\tilde{d}_s$ is monotonic, we are allowed
to reason about the marginal improvements by comparing their derivatives, which is:
\begin{equation}
    \frac{\partial (\frac{1}{\tilde{d}_s} - \frac{1}{d_s})}{\partial \tilde{d}_s} = -\frac{1}{\tilde{d}_s^2}
\end{equation}
Notice that the derivative does not depend on the degree $d_s$ of the vertex $s$ at all,
and the greater the magnitude of the derivative, the more improvement the variance of a
vertex gets by sampling one more edge. Thus, choosing any vertex $s$ with least number
of edges sampled would work for us, that is: $s = \argmin_{s' \in S} \tilde{d}_{s'}$.
In light of this observation, one can see that it is optimal to sample an equal number
of edges for each vertex in $S$. This is one of the reasons LADIES is not efficient with
respect to the number of edges it samples.
On graphs with skewed
degree distributions, it samples thousands of edges for some seed vertices, which
contribute very small amounts to the variance of the estimator since it is already
very close to $0$.

\subsection{Sampling a fixed number of neighbors}
\label{subsecl:fixed_neighbors}

One can easily resort to Sequential Poisson Sampling by~\cite{Ohlsson1998} if one wants $\tilde{d}_s = \min(k, d_s)$ instead of $E[\tilde{d}_s] = \min(k, d_s)$ to get the exact
same behavior to Neighbor Sampling. Given $\pi_t$, $c_s$ and $r_t$, we pick the $\tilde{d}_s = \min(k, d_s)$ smallest vertices $t \to s$ with respect
to $\frac{r_t}{c_s \pi_t}$, which can be computed in expected linear time by using the quickselect algorithm~\citep{Hoare1961}.

\subsection{Baseline Comparison of Vertex/Edge Efficiency (cont.)}
\label{subsecl:same_convergence}

\begin{figure*}[!h]
    \centering
    \includegraphics[width=\linewidth]{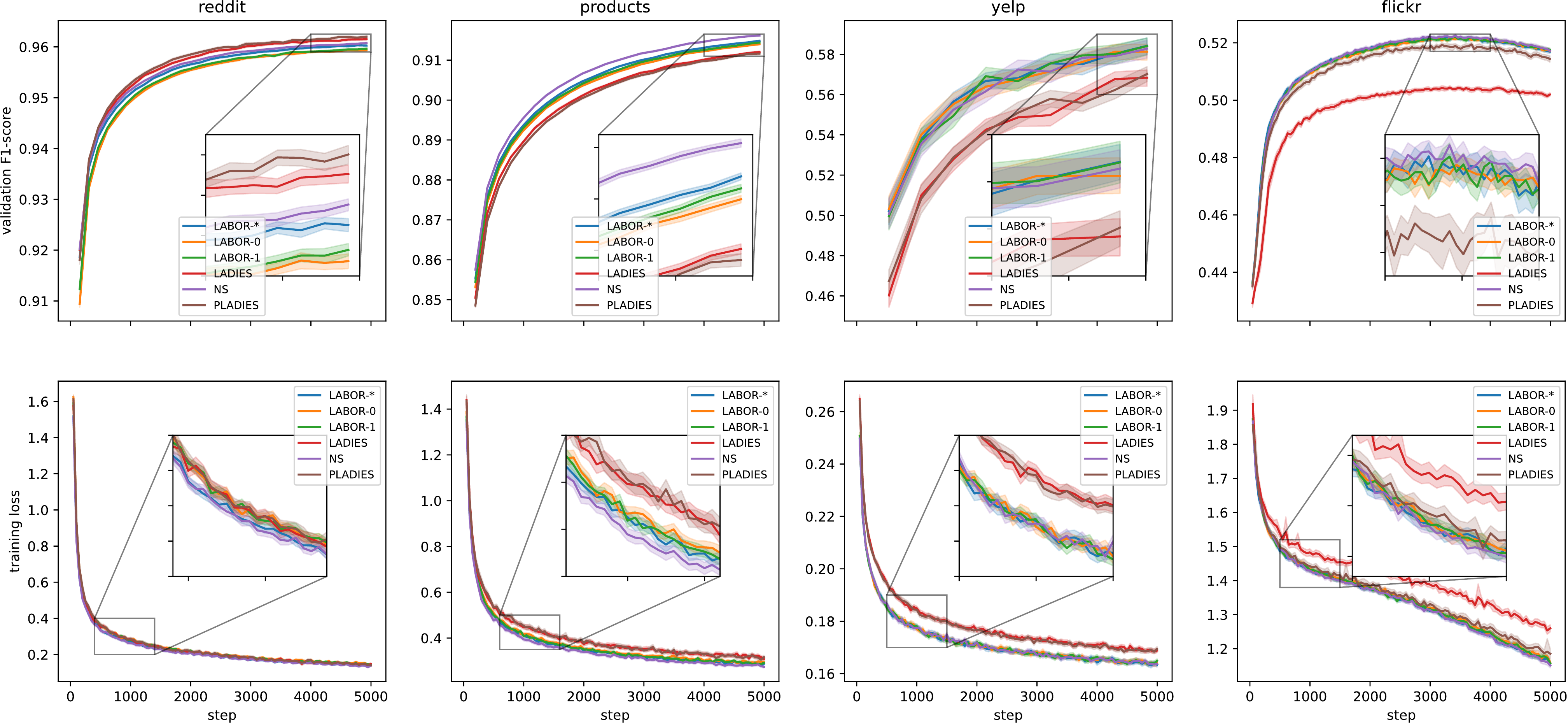}
    \caption{The validation F1-score and training loss curves on different datasets with same batch size. The soft edges represent the confidence interval. The x-axis stands for the number of training iterations. Number of sampled vertices and edges can be found in~\cref{tabl:num_sampled}.}
    \label{figl:same_loss}
\end{figure*}

Here, we show another version of~\cref{figl:same_loss_cum_both}, replacing the x-axis with 
the number of training iterations, see~\cref{figl:same_loss}. We would like to point out that despite NS and 
LABOR variants sampling drastically different number of vertices/edges as shown
in~\cref{tabl:num_sampled} (up to $7\times$ fewer vertices), their
convergence behavior is more or less the same, even indistinguishable on Yelp
and Flickr.

\subsection{Fixed point iterations}
\label{subsecl:fixed_point_iterations_proof}

Given any $\pi^{(0)} > 0$ and one iteration to get $\pi^{(1)}$ and one more iteration using $c(\pi^{(1)})$ to get $\pi^{(2)}$, then we have the following observation:

\begin{gather*}
    \pi^{(1)}_t = \pi^{(0)}_t \max_{t \to s} c_s(\pi^{(0)}) \\
    \frac{1}{d_s^2}\sum_{t \to s} \frac{1}{\min(1, c_s(\pi^{(0)}) \pi^{(0)}_t)} - \frac{1}{d_s} = \frac{1}{k} - \frac{1}{d_s} \\
    \frac{1}{d_s^2}\sum_{t \to s} \frac{1}{\min(1, c_s(\pi^{(1)}) \pi^{(1)}_t)} = \frac{1}{d_s^2}\sum_{t \to s} \frac{1}{\min(1, c_s(\pi^{(1)}) \max_{t \to s'} c_{s'}(\pi^{(0)}) \pi^{(0)}_t)}
\end{gather*}

Now, note that for a given $t \in N(s)$, $\max_{t \to s'} c_{s'}(\pi^{(0)}) \geq c_s(\pi^{(0)})$ since $s \in \{s' \mid t \to s'\}$. This implies that $\max_{t \to s'} c_{s'}(\pi^{(0)}) \pi^{(0)}_t \geq c_s(\pi^{(0)}) \pi^{(0)}_t$. Note that for $\pi_t' = c_s(\pi^{(0)}) \pi^{(0)}_t, \forall t \to s$, we have $c_s(\pi') = 1$ for any given $s$.
Since $\max_{t \to s'} c_{s'}(\pi^{(0)}) \pi^{(0)}_t \geq c_s(\pi^{(0)}) \pi^{(0)}_t = \pi_t', \forall t \to s$, this let's us conclude that 
the $c_s(\pi^{(1)}) \leq 1$, because the expression is monotonically increasing with respect to any of the $\pi_t$. By induction, this means 
that $c_s^{(i)} \leq 1, \forall i \geq 1$. Since $\pi^{(i)}_t = \pi^{(0)}_t \prod_{j = 0}^{i - 1}\max_{t \to s'} c_{s'}(\pi^{(i - 1)})$,
$\pi^{(i)}_t$ is monotonically decreasing. This means that the objective value
in~(\ref{eq:objective_function}) is also monotonically decreasing and is clearly bounded from below by 
$0$. Any monotonically decreasing sequence bounded from below has to converge,
so our fixed point iteration procedure is convergent as well.

The intuition behind the proof above is that after updating $\pi$
via~(\ref{eq:pi_fixed_point_iteration}), the probability of each edge 
$t \to s: \pi_t c_s$ goes up because the update takes the maximum $c_s$ over 
all possible $s$. This means each vertex gets higher quality estimators than the
set variance target so we now have room to reduce the number of vertices sampled by 
choosing the appropriate $c_s \leq 1$.
To summarize, $\pi$ update step increases the quality of the batch which in turn let's the
$c_s$ step to reduce the total number of vertices sampled.

\subsection{GATv2 runtime performance}

We utilize the GATv2 model~\cite{brody2022how} and show the runtime performance of different samplers in~\cref{tabl:gat_runtime}. All the hyperparameters were kept same as~\cref{subsecl:NS_comp} and the number of attention heads was set to 8. The runtimes correlate with the $|E_3|$ column in~\cref{tabl:num_sampled} for the GATv2 model as the computation and memory requirements depend on the number of edges. That is why the LADIES variants went out-of-memory on the reddit and products datasets.

\begin{table}
\centering
\begin{small}
\caption{The runtimes (ms) per iteration for the GATv2 model on NVIDIA A100 80GB corresponding to~\cref{tabl:num_sampled}. The abbreviation OOM stands for an ‘out-of-memory’ error, which occurs when an excessive number of edges are sampled.}
\label{tabl:gat_runtime}
\begin{tabular}{c | r | r | r | r | r | r}
\toprule
\textbf{Dataset} & \textbf{LADIES} & \textbf{PLADIES} & \textbf{LABOR-*} & \textbf{LABOR-1} & \textbf{LABOR-0} & \textbf{NS} \\
\midrule
reddit & \color{Maroon}{\textbf{OOM}} & \color{Maroon}{\textbf{OOM}} & 266.6 & 69.8 & \color{OliveGreen}{\textbf{51.9}} & 155.1 \\
products & \color{Maroon}{\textbf{OOM}} & \color{Maroon}{\textbf{OOM}} & 224.8 & 154.6 & \color{OliveGreen}{\textbf{138.7}} & 218.5 \\
yelp & \color{Maroon}{\textbf{278.0}} & \color{Maroon}{\textbf{278.9}} & 169.2 & 96.6 & \color{OliveGreen}{\textbf{83.2}} & 99.5 \\
flickr & \color{Maroon}{\textbf{96.6}} & \color{Maroon}{\textbf{100.3}} & 79.8 & 65.0 & \color{OliveGreen}{\textbf{61.1}} & 66.2 \\
\bottomrule
\end{tabular}
\end{small}
\end{table}

\subsection{Extension to the weighted case}
\label{subsecl:weighted_extension}

If the given adjacency matrix $A$ has nonuniform weights, then we want the estimate the following:

\begin{equation}
    H_s = \frac{1}{A_{*s}}\sum_{t \to s} A_{ts} M_t
\end{equation}

where $A_{*s} = \sum_{t \to s} A_{ts}$. If we have a probability over each edge $\pi_{ts}, \forall (t \to s) \in E$, then the variance becomes:

\begin{equation}
    \Var(H''_s) \leq \Var(n H'_s) = \frac{1}{(A_{*s})^2} \Big( \sum_{t \to s} \frac{A_{ts}^2}{\min(1, c_s \pi_{ts})} - \sum_{t \to s} A_{ts}^2 \Big)
\end{equation}

In this case, we can still aim to reach the same variance target $v_s = \frac{1}{k} - \frac{1}{d_s}$ or any given custom target $v_s \in \mathbb{R^+}$ by finding $c_s$ that satisfies the following equality:

\begin{equation}
    \frac{1}{(A_{*s})^2} \Big( \sum_{t \to s} \frac{A_{ts}^2}{\min(1, c_s \pi_{ts})} - \sum_{t \to s} A_{ts}^2 \Big) = v_s
\end{equation}

In this case, the objective function becomes:

\begin{equation}
     \label{eq:weighted_objective_function}
     \pi^* = \argmin_{\pi \geq 0} \sum_{t \in N(S)} \min(1, \max_{t \to s} c_s \pi_{ts})
\end{equation}

Optimizing the objective function above will result into minimizing the number of vertices sampled. 
Given any $\pi_{ts} > 0, \forall (t \to s)$,
then the fixed point iterations proposed for the non-weighted case
in~(\ref{eq:pi_fixed_point_iteration}) can be modified as follows:

\begin{equation}
    \pi^{(0)} = A, \forall (t \to s): \pi^{(i + 1)}_{ts} = \max_{t \to s'} c_{s'}(\pi^{(i)}) \pi_{ts'}^{(i)}
    \label{eq:pi_weighted_fixed_point_iteration}
\end{equation}

A more principled way to choose $v_s$ in the weighted case is by following the argument presented
in~\cref{subsecl:choosing_how_many}. There, the discussion revolves
around the derivative of the variance with respect to the expected number of vertices sampled for a given seed vertex $s$. If we apply the same argument, then we get:

\begin{gather*}
    v_s(c_s) = \frac{1}{(A_{*s})^2} \Big( \sum_{t \to s} \frac{A_{ts}^2}{\min(1, c_s \pi_{ts})} - \sum_{t \to s} A_{ts}^2 \Big) \\
    \frac{\partial v_s}{\partial c_s} = \frac{-1}{(A_{*s})^2} \sum_{t \to s} \frac{\mathbbm{1}[c_s \pi_{ts} < 1] A_{ts}^2}{c_s^2 \pi_{ts}} \\
    E[\tilde{d}_s](c_s) = \sum_{t \to s} \min(1, c_s \pi_{ts}) \\
    \frac{\partial E[\tilde{d}_s]}{c_s} = \sum_{t \to s} \mathbbm{1}[c_s \pi_{ts} < 1] \pi_{ts}
\end{gather*}

Then to compute the derivative of the variance with respect to the expected number of vertices sampled for a given seed vertex $s$,
which is $\frac{\partial v_s}{\partial E[\tilde{d}_s]}$, we can use the chain rule and get:

\begin{equation}
    \frac{\partial v_s}{\partial E[\tilde{d}_s]} = \frac{\partial v_s}{\partial c_s}\frac{\partial c_s}{\partial E[\tilde{d}_s]} = \frac{\frac{-1}{(A_{*s})^2} \sum_{t \to s} \frac{\mathbbm{1}[c_s \pi_{ts} < 1] A_{ts}^2}{c_s^2 \pi_{ts}}}{\sum_{t \to s} \mathbbm{1}[c_s \pi_{ts} < 1] \pi_{ts}}
\end{equation}

Then, what one would do is to choose a constant $C(k)$ as a function of the fanout parameter $k$ and set $\frac{\partial v_s}{\partial E[\tilde{d}_s]} = C(k)$ and solve for $c_s$.
$C(k)$ would be a negative quantity whose absolute value decreases as $k$ increases. It would probably look like $C(k) = \frac{-C'}{k^2}$ for some constant $C' > 0$. We leave this
as future work.

\subsection{Convergence speed with respect to wall time}

In this section, we perform hyperparameter optimization on Neighbor Sampler and Labor Sampler so that the training converges to a target validation accuracy as fast as possible.
We leave LADIES out of this experiment because it is too slow as can be seen in the last column of~\cref{tabl:num_sampled}.
We ran this experiment on an A100 GPU and stored the input features on the main memory, which were accessed over the PCI-e directly during training by pinning their memory.
This kind of training scenario is commonly used when training on large datasets whose input features don't fit in the GPU memory.
We use larger of the two datasets we have for this experiment, products and yelp.
For products, the validation accuracy target we set is 91.5\%, and for yelp it is 60\%. We tune the learning rate between
$[10^{-4}, 10^{-1}]$, the batch size between $[2^{10}, 2^{15}]$ and the fanout for each layer between $[5, 25]$. We use the same model used in~\cref{secl:experiments}. For LABOR,
we additionally tune over the number of importance sampling iterations i between $[0, 3]$ so that it can switch between LABOR-$i$ and also a layer dependency boolean parameter that
makes LABOR use the same random variates $r_t$ for different layers when enabled, which has the effect of increasing the overlap of sampled vertices across layers. We use the state
of the art hyperparameter tuner HEBO~\cite{cowen2020empirical}, the winning submission to the NeurIPS 2020 Black-Box Optimization Challenge, to tune the parameters of the sampling 
algorithm with respect to runtime required to reach the target validation accuracy with a timeout of 300 seconds, terminating the run if the configuration doesn't reach the target
accuracy.

We let HEBO run overnight and collect the minimum runtimes required to achieve the target accuracies. For products,
the fastest hyperparameter corresponding to the 38.2s runtime had fanouts $(18, 5, 25)$, batch size $10500$, learning rate $0.0145$, used LABOR-1, layer dependency False.

For Neighbor Sampler, the fastest hyperparameter corresponding to 43.82s runtime had fanouts $(15,5,21)$, batch size $12000$, learning rate $0.0144$.

For the Yelp dataset, the fastest hyperparameter corresponding
to the 41.60s runtime had fanouts $(6, 5, 7)$, batch size $5400$, learning rate $0.000748$, used LABOR-1 and layer dependency True.

For Neighbor Sampler, the fastest hyperparameter corresponding to
the 47.40s runtime had fanouts $(5, 6, 6)$, batch size $4600$, learning rate $0.000931$.

\begin{figure}[h]
    \centering
    \includegraphics[width=0.70\linewidth]{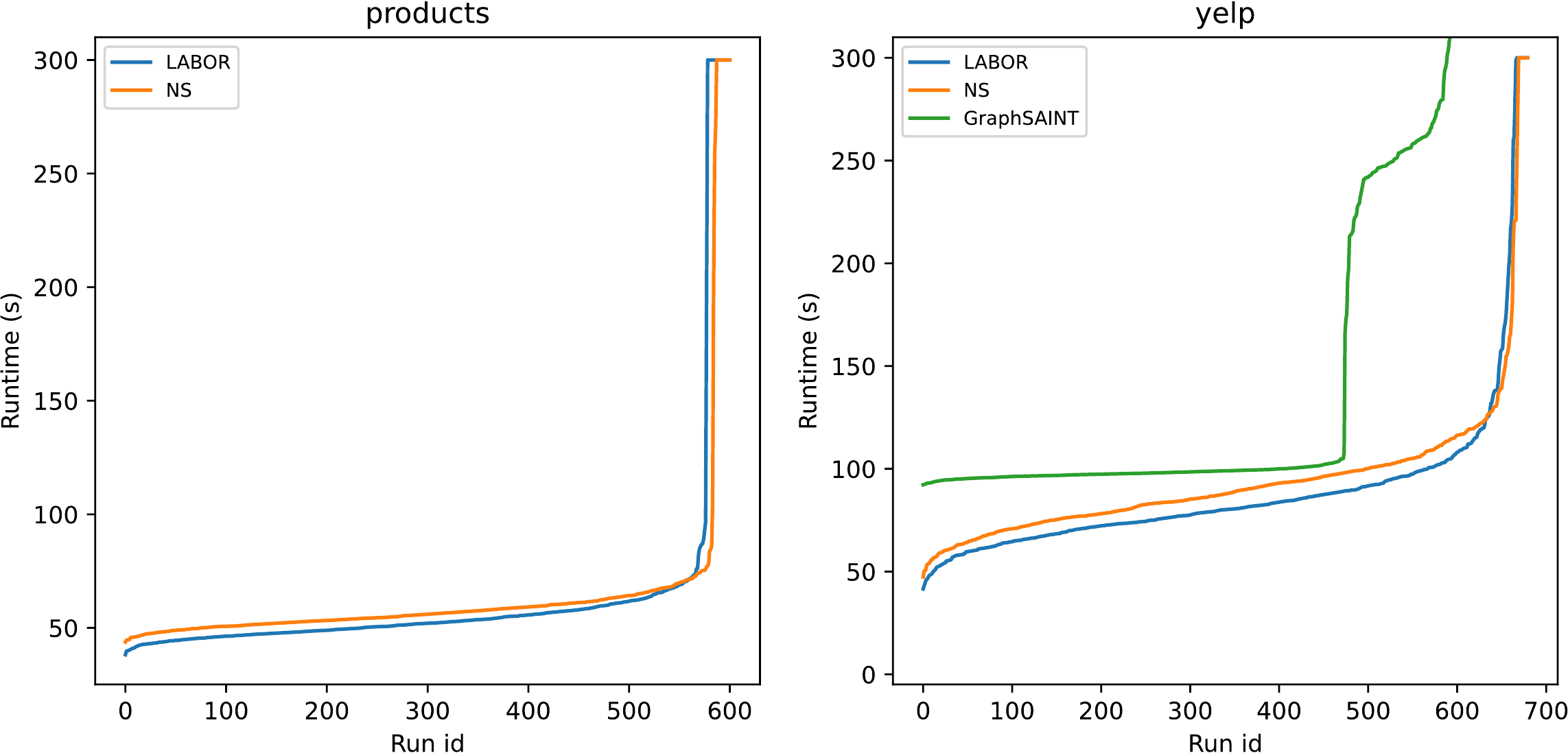}
    \caption{The runtimes to reach a validation F1-score of 91.5\% on products and 60\% on yelp, belonging to runs tried by the HEBO hyperparameter tuner, sorted with respect to their runtimes. Having a lower curve means that a method is faster overall compared to the others. HEBO could not find any hyperparameter
    configuration to reach the set target on products, hence its curve is left out.}
    \label{figl:hebo_sorted}
\end{figure}

These results clearly indicate that training with LABOR is faster compared to Neighbor Sampling when it comes to time to convergence.

We run the same experiment with GraphSAINT~\citep{graphsaint-iclr20} using the DGL example code, both on ogbn-products and yelp using the same model architecture. We used their edge sampler, the edge sampling budget between
$[2^{10}, 2^{15}]$ and learning rate between $[10^{-4}, 10^{-1}]$. We disabled batch normalization to make it a fair comparison with NS and LABOR since the models
they are using does not have batch normalization. We use HEBO to tune these hyperparameters to reach the set validation accuracy, and let it run overnight.
The results show that HEBO was not able to find any configuration of the hyperparameters reaching 91.5\% accuracy on products faster than 1500 seconds. For the Yelp
dataset, the fastest runtime to reach 60\% accuracy was 92.2s, with an edge sampling budget of $12500$ and learning rate $0.0214$.

\end{document}